**Applied Computational Intelligence and Soft Computing**

# Lion Algorithm- Optimized Long Short-Term Memory Network for Groundwater Level Forecasting in Udupi District, India

Supreetha B.S[1], Narayan Shenoy[2] and Prabhakar Nayak [3]

[1, 3] Department of ECE, Manipal Institute of Technology, Manipal, 576104, India
[2] Department of Civil Engineering, Manipal Institute of Technology, Manipal, 576104, India

Corresponding Author: Supreetha B.S (supreetha.bs@manipal.edu)

## Abstract

Groundwater is a precious natural resource. Groundwater level (GWL) forecasting is crucial in the field of water resource management. Measurement of GWL from observation-wells is the principle source of information about the aquifer and is critical to its evaluation. Most part of the Udupi district of Karnataka State in India consists of geological formations: lateritic terrain and gneissic complex. Due to the topographical ruggedness and inconsistency in rainfall, the GWL in Udupi region is declining continually and most of the open wells are drying-up during the summer. Hence, the current research aimed at developing a groundwater level forecasting model by using hybrid Long Short-term Memory-Lion Algorithm (LSTM-LA). The historical GWL and rainfall data from an observation well from Udupi district, located in Karnataka state, India, were used to develop the model. The prediction accuracy of the hybrid LSTM-LA model was better than that of the Feedforward Neural network (FFNN) and the isolated LSTM models. The hybrid LSTM-LA based forecasting model is promising for a larger dataset.

## 1. Introduction

The ground water (GW) survey reported in The New Indian Express in 2018 reveals that groundwater level (GWL) is very poor in South Indian states. The survey reports that out of 1421 wells surveyed in Karnataka, 985 showed a decline in GWLs. The survey also reported that GWL is declining continuously in Udupi district [1]. It is crucial to be able to forecast GW resources with an appropriate model using advanced algorithms. GWL forecasting system has more than six decades of history [2]. A large number of research work is found in the literature, which has reached a certain level of maturity.

The hydrogeological GWL forecasting models are probabilistic, deterministic and stochastic for the assessment of GW systems. The traditional GW flow models are partial differential equations, which are embedded with simplifying assumptions about the aquifer properties and boundary conditions [3]. These natural groundwater systems are complex and have a large number of parameters that are highly variable throughout time and space, such as aquifer parameters like hydraulic conductivity of the formation, groundwater storage, dimension of the aquifer and other parameters related to the geological structure. To simplify GWL forecasting, researchers have tried to explore various parameters to develop GWL forecasting models [4].

The importance of the hydrological model for environment and water management is growing with urbanization and climate variability. The hydrological models are broadly classified into conceptual, physical and mathematical models [5]. Mathematical models are further categorized as empirical lumped conceptual and physically-based models. Physically based models use physically measurable static input variables and require extensive information about the study area. Measuring physical properties is difficult, especially for predictive models, where the input values change over time [6]. Physical models though accurate in prediction, are not very practical as they are less efficient in predicting irregularly varying patterns of data [7]. To overcome this limitation and with the rapid increase in computation power, recently data-driven models are adopted using quantitative historical data to forecast future trends [8], which have become a standard tool in water resources management sector [9]. Machine learning-based approaches are promising for hydrological time series forecasting. However, many of the techniques rely on optimization of Artificial Neural Network (ANN) weights or architectures. Data driven models are developed with existing data and information on the relationship between input and output parameters. These models are location specific, with the output values being applicable only to the location where it is developed [10]. Statistical, fuzzy, regression and ANN are mathematical approaches typically used in these data-driven models. ANN models have received much interest in recent literature [11]. Researchers have implemented the functionality of ANNs to model surface and groundwater quantity [12]. Backpropagations (BP) are extensively used for ANN training. However, the results of ANN approach are found to be less consistent and unstable [13]. Hence, alternative and advanced data-driven models are required for predicting real-time GWL more accurately.

Different types of ANN architectures and algorithms are developed in the literature using multilayer feedforward, recurrent networks and radial basis networks [14]. Sethi et al. [15] investigated multilayer feedforward with BP learning algorithm to develop water table depth forecasting model. Exploring the important parameters that influence the water table fluctuations, they employed monthly rainfall, evapotranspiration and water table depth as input parameters. They predicted groundwater table depth for one month ahead in a hard rock aquifer. The models were calibrated with limited input dataset monitored during study period. The performance of the model can further be improved with sufficient datasets and with different architectures. The traditional ANNs cannot handle sequential data effectively, which is one of the major drawbacks [16]. Predictive models with longer lead-time are required which have been developed using deep learning techniques with multiple hidden layers.

Deep learning techniques with multiple hidden layers between the input and Recurrent Neural Networks (RNN) are widely used in recent years [17, 18]. However, the standard RNN architecture has difficulty in capturing long-term dependencies between variables, due to vanishing and exploding gradient problem, which can be overcome by a variant of RNN called Long Short Term Memory (LSTM). LSTM has only recently been used for hydrologic time series prediction [19]. Bowes et al. [20] compared RNN with LSTM for predicting GW table in the flood-prone coastal city of Norfolk, Virginia. They explored two machine learning algorithms LSTM and RNN to model and predict GW table response to storm events, using GW table, rainfall and sea level as input parameters from 2010-2018 to train and the models. As per their study LSTM networks were found to have more predictive skills than RNN's. Kratzert et al. [21] explored application of LSTM as a regional rainfall-runoff model in catchments of the freely available CAMELS dataset. They tested their approach and compared the results to the well-known Sacramento Soil Moisture Accounting Model (SAC-SMA) and achieved better model performance, which underlined the potential of LSTM for hydrological modelling applications. The LSTM RNN has an internal state and may learn to forecast different series with good long-term memory, which is one of the most attractive and powerful features compared to traditional Feedforward Neural Network (FFNN).

There are several drawbacks for using an LSTM network in isolation. Learning LSTM models for large number of memory cells becomes computationally expensive. It also suffers from the lack of ability to explain the final decision that the model acquires [22]. To overcome this limitation, hybrid approach has been used. Nawi et al. [23] investigated the data classifier problem by employing weight optimization on RNN using cuckoo search hybrid techniques. The convergence rate and local minima problem are addressed as cuckoo search algorithm. The performance of this model is compared with ABC using BPNN algorithm and other hybrid variants. The results show that the computational efficiency of traditional RNN is highly improved when coupled with the hybrid method. Chung et al. [24] investigated a novel stock market prediction model using the available financial data. They adopted the deep learning technique of hybrid approach by integrating LSTM with a genetic algorithm. They used a systematic method to determine the time window size and topology of the LSTM network using Genetic Algorithm (GA). The experimental results demonstrated that the hybrid LSTM network outperforms the benchmark model. Rashid et al. [25] developed a well-structured LSTM for resolving difficulties with traditional RNN networks. They used four different optimizers based on metaheuristic algorithms, Harmony Search (HS), Gray Wolf Optimizer (GWO), Sine Cosine Algorithm (SCA) and Ant Lion Optimization Algorithm (ALOA). The learning speed and accuracy due to long-term dependencies in LSTM are explored and compared with RNN architecture. They suggested that the classification accuracy of LSTM outperforms traditional RNN architecture and the increased complexity in training these networks could be resolved by using alternative, powerful, nature inspired algorithms.

There is a need to have a computationally efficient model that can forecast water levels with minimum parametrization. At the same time, such a model should be able to deal with expected climate variability. To overcome the weakness and to improve the convergence rate (prediction accuracy) of traditional approaches more advanced, simple, robust, efficient and accurate model is required. Lion algorithm (LA) is a nature-inspired algorithm developed by B.R.Rajkumar in 2012, which mimics social territorial lions breeding and its defence to other nomadic lions. This LA can be used in conjunction with LSTM to find the optimal solutions. The current study aims at developing a new hybrid metaheuristic approach using the LA to optimise the weights of LSTM network. The study also aims analyse the performance of the proposed hybrid LSTM-LA approach on a selected dataset by comparing with the standard feedforward architecture.

## 2. Materials and Methods

The study developed and tested a hybrid LSTM-LA model. This section describes the study area and the dataset used, description of FFNN model, architecture of the proposed model and its implementation.

**2.1 Study area and dataset**: One of the challenges in GWL forecasting is that the flow of groundwater is unique to geological formations. Therefore, the GW analysis is site region-specific. No standard benchmark can be used for the forecasting of GWL to build the predominance of the model. It is therefore essential to develop regional GWL forecasting by collecting data from the specific region. The study is based on the secondary data from government agencies collected from an observation-well located in Udupi district of Karnataka state in India (Figure 1). The geology of the Karnataka state is very complex with varied parameter in its formations from laterites, gneisses granites, dolerite dykes and coastal sedimentary rock types. [26].The observation well considered in this study is located in lateritic terrain [27].

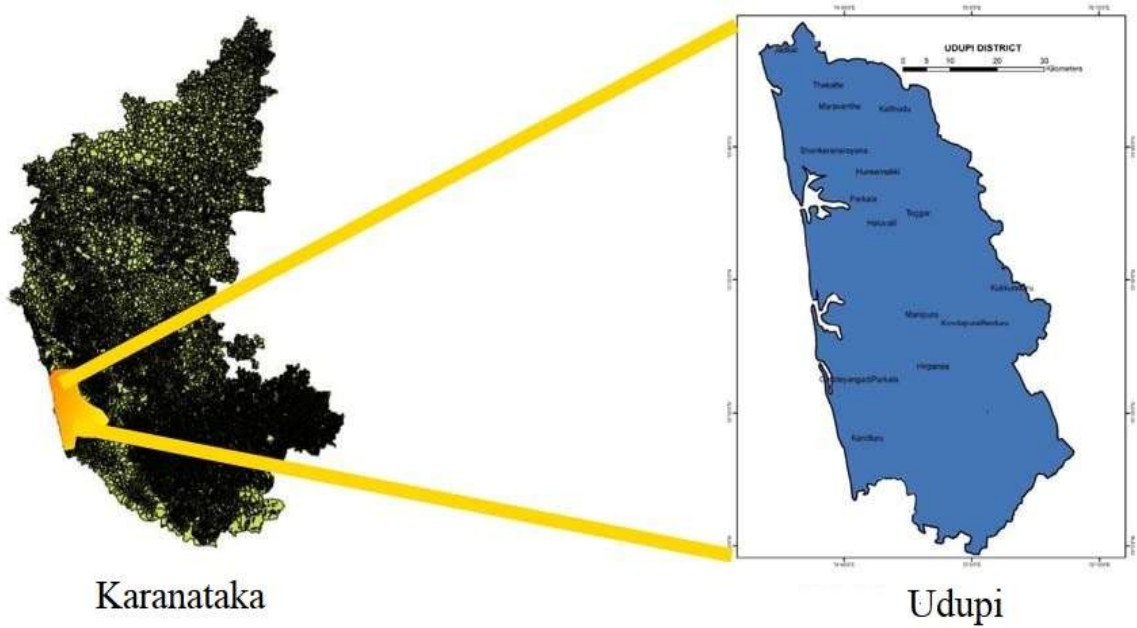

Figure 1:  Location map of Udupi district

## 2.2 Feedforward Neural Network based groundwater level forecasting approach:

The FFNN structure for forecasting GWL has wide application in the GW studies. The most frequently used algorithm for aquifer models in neural network domain is gradient descent algorithm. In this work, the weights of the FFNN were optimised using gradient descent approach. The conventional gradient descent based algorithms operate on a single weight vector. The FFNN structure with two input, three hidden and one output node with gradient descent training as shown below (Figure 2).

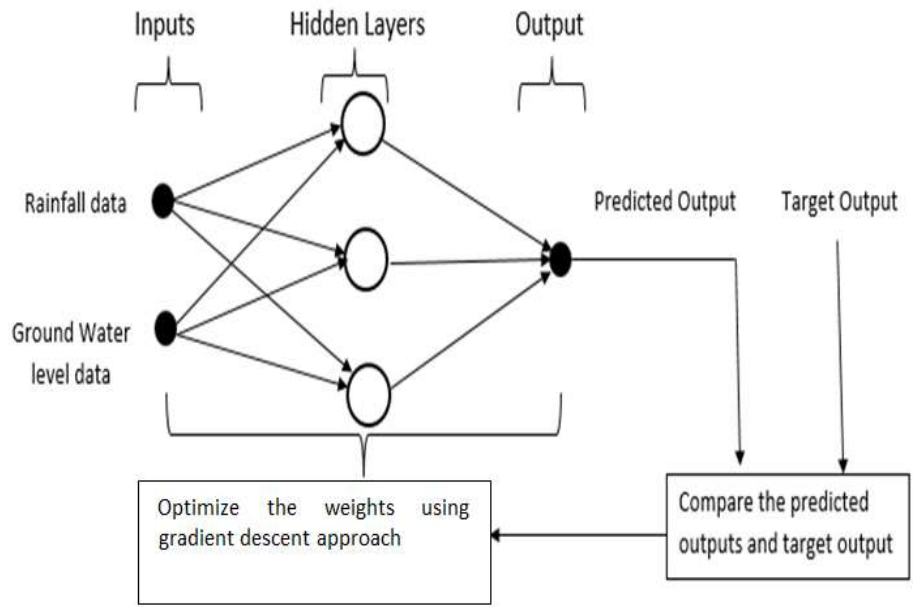

Figure 2: The FFNN weight optimization using gradient descent approach

The FFNN configurations learn in a randomised order and the information only flows in the forward direction in every layer of the network. Since there is no looping, it predicts only continuous target variables. Therefore, in order to learn progressively deep learning algorithm, special type of RNN called LSTM approach with self-connected gates in the hidden layer is implemented.

**2.3 Hybrid Long Short Term Memory-Lion Algorithm (LSTM-LA) approach:** LSTMs, introduced by Hochreiter and Schmidhuber (1997), are special kind of RNNs capable of learning long-term dependencies. LSTMs selectively remember patterns for long durations of time compared to traditional FFNN. LSTMs, capable of removing or adding information to the cell state through carefully regulated gates such as forget gate f, input gate i, input modulation gate g and output gate (Figure 3).

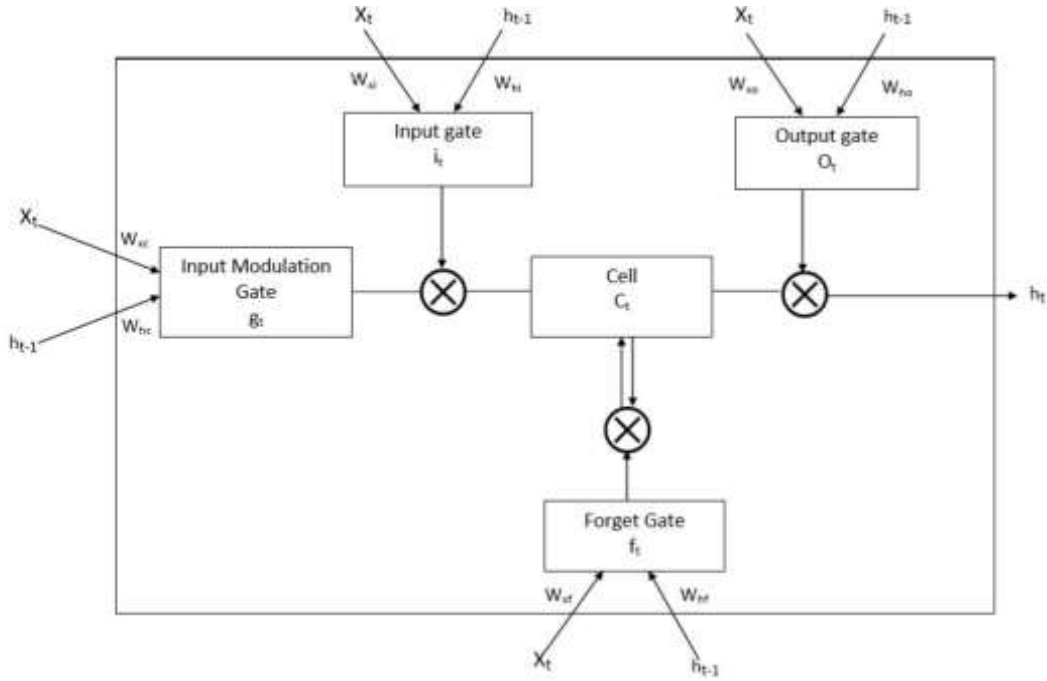

Figure 3: Structure of LSTM gates

The forget gate helps to process the output of previous state $h_{t-1}$, to take decisions by forgetting unnecessary information. The forget layer with sigmoid function is represented in Equation 1. The input gate adds new information with appropriate scaling and sigmoid activation function updates the values and tanh function creates new candidate values (Equation 2 and Equation 3). The updated new candidate value with proper scaling is also given in Equation 4.

$$f_t = \sigma(W_f . [h_{t-1}, x_t] + b_f) \qquad (1)$$

$$i_t = \sigma( W_i . [h_{t-1}, x_t] + b_i \qquad (2)$$

$$\acute{C}_t = tanh (W_C . [h_{t-1}, x_t] + b_C \qquad (3)$$

$$C_t = f_t * C_{t-1} + i_t * \acute{C}_t \qquad (4)$$

Finally, the relevant output of sigmoid function is represented in Equations 5 and 6.

$$o_t = \sigma (W_o {[h_{t-1,} x_t]} + b_o) \qquad (5)$$

$$h_t = o_t * tanh (C_t) \qquad (6)$$

The basic LSTM neuron has a separate cell state that keeps track of long-term sequential information. However, learning LSTM models for large number of memory cell becomes computationally expensive. Therefore, a hybrid LSTM-LA methodology is adopted in the current study as shown by the flow diagram (Figure 4).

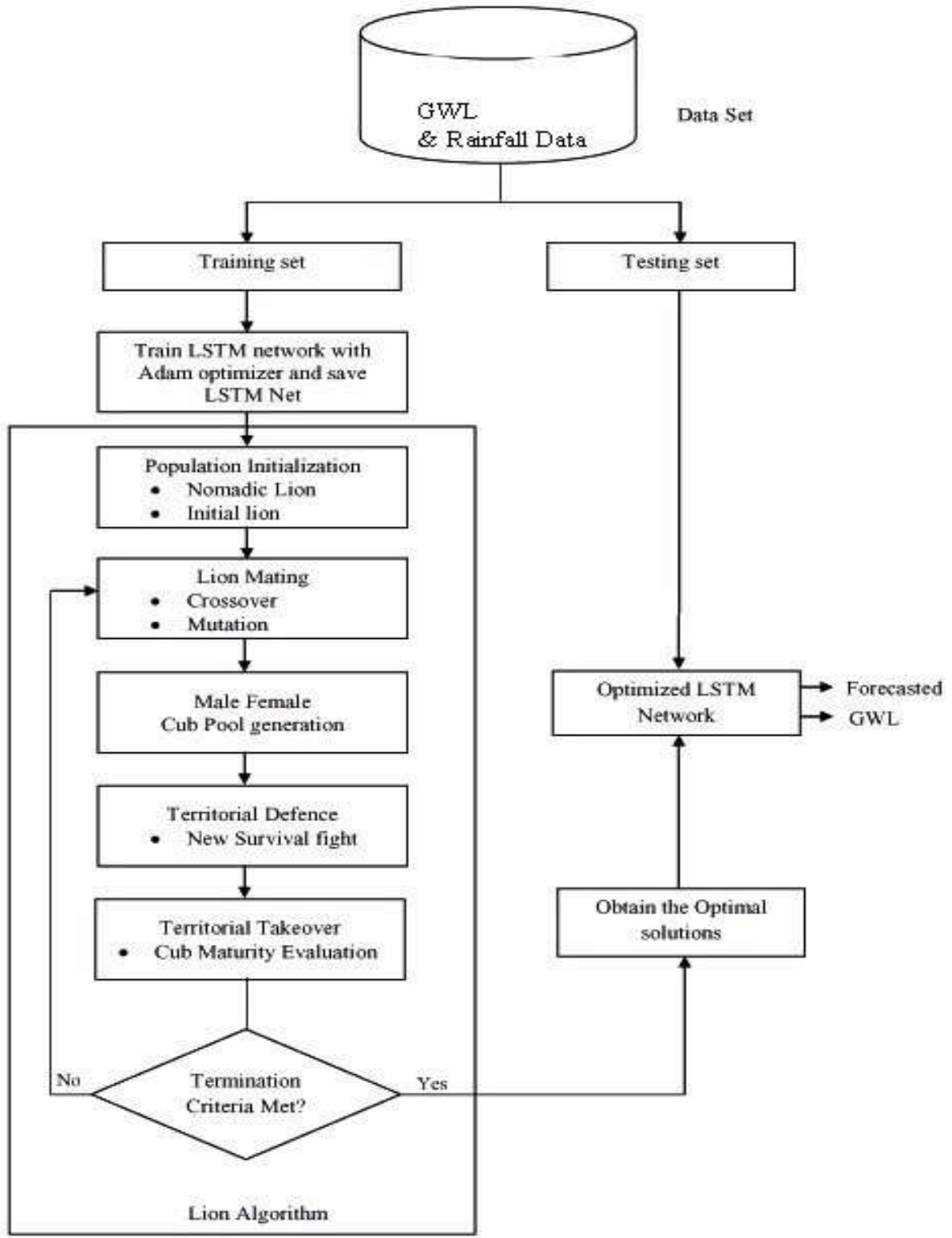

Figure 4: Flow diagram representing the hybrid LSTM-LA methodology

In the hybrid LSTM-LA model, the mating characteristics of lions are mathematically modelled to optimise the weights of LSTM network. The population of randomly generated set of solutions called lions are initialised. The possible solutions are the weights and biases for LSTM network. The population of 2n lions are assigned to two groups as the candidate population. The best weights and biases are initialised with LA in the first epoch and are passed on to LSTM network. The second step in the algorithm is the mating process that assures the lion's survival as well as a platform for information exchange among different members. The new cubs are produced after selecting the female and male lions using linear combination of parents using mating operators as given below in Equations 7 and 8.

$$Offspring_j 1 = \alpha \times Female\ Lion_j + \sum \frac{1-\alpha}{\sum_{i=1}^{NRM} Si} \times Male\ Lion_{j}{}^{i} \times S_i \quad (7)$$

$$Offspring_j 2 = (1-\alpha) \times Female\ Lion_j + \sum \frac{\alpha}{\sum_{i=1}^{NRM} Si} \times Male\ Lion_{j}{}^{i} \times S_i \quad (8)$$

Here, NRM = number of residents males in a pride, $\alpha$ = randomly generated number and

$$S_i = \begin{cases} 1, & If\ male\ i\ is\ selected\ for\ mating \\ 0, & otherwise \end{cases}$$

The mutation operator with mutation rate of 0.2 is applied randomly on each gene of the offspring. The last stage in LA is defence operator, which consists of defence against new matured resident males and defence against nomad lions. This defence operator plays an important role in LA by assisting it to retain powerful male lions as solutions. The nomadic lion is generated in the same way as territorial lion and new survival fight between territorial lion and nomadic lion is performed. The male lion occupies the territory by defending and protecting the cubs and then the new solution is used to attack the male lion. If the nomadic lion is superior to the other solutions in the pride, the male lions are replaced by the nomadic lion. The territorial takeover is the last step, which is the same as selection process in genetic algorithm. In this step, the optimal solution found to replace the inferior one and mating process is repeated until termination condition of 100 epochs are reached. The LA will update weights with best possible solutions in the next cycle and the searching process is continued.

Thus, the weights and thresholds of all layers in the LSTM model are initialized randomly and LA searches the optimal weights. If the termination criteria, i.e. the maximum iteration number is reached, the optimal parameters are obtained; or else the optimization steps are repeated until the conditions are satisfied. Then, the optimized LSTM model is used to forecast the GWL.

The hybrid LSTM-LA, LSTM and FFNN models are used to forecast the future trend of GWL. The dataset from the period year 2000-2018 was used to train and test the LSTM-LA model for different prediction horizon. The 80% of the data is set as training set and remaining 20% is set as testing set. The monthly forecast of GWL results for year 2018 from the hybrid LSTM-LA model are compared with LSTM and FFNN approaches.

## 3. Results and Discussion

The GWL forecast for year 2018 using the hybrid LSTM-LA, LSTM and FFNN approaches are shown below (Figure 5). The forecast results were verified against observed results. The FFNN based prediction shows large error in premonsoon period compared to hybrid LSTM-LA approach. It was observed that the hybrid LSTM-LA model results are correlating better with the original data compared to the LSTM and FFNN other approaches. Thus, LSTM-LA approach predicts more accurately compared to traditional feedforward approach.

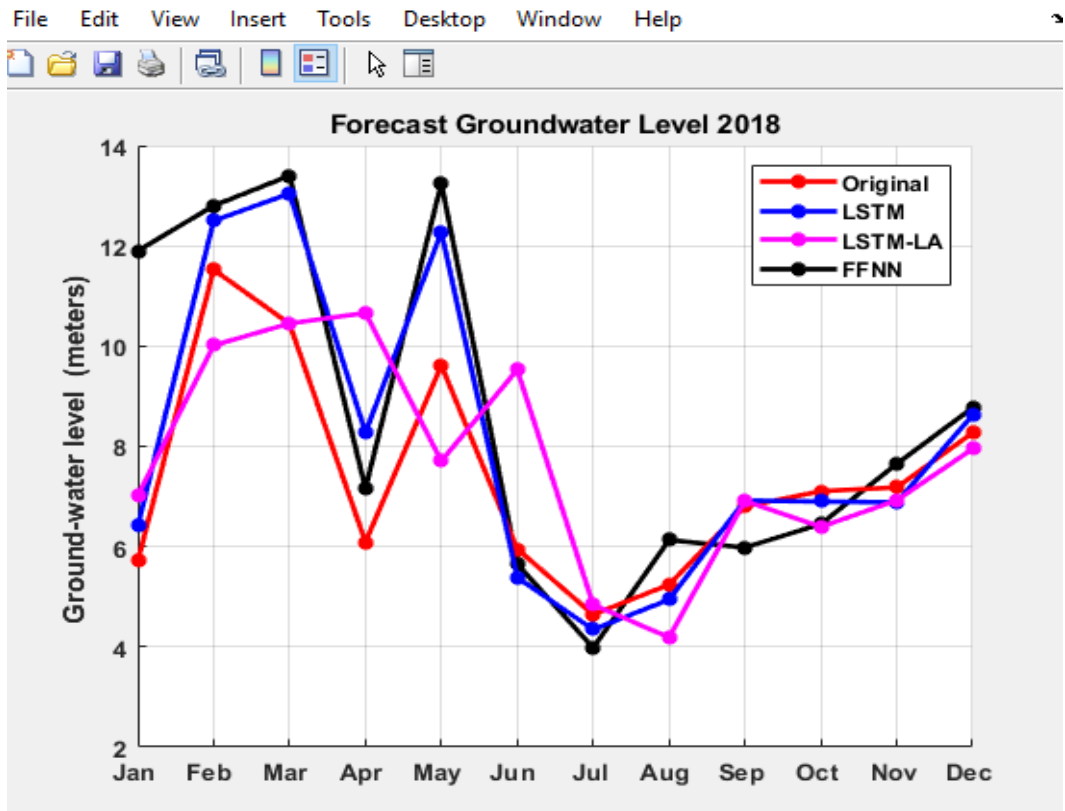

Figure 5: Comparative plot of monthly forecast GWL using FFNN, LSTM and LSTM-LA

We considered two performance metrics to assess the forecasting accuracy. Figure 6 shows the performance of all the three soft computing approaches using statistical indices Root Mean Square Error (RMSE) and Mean Absolute Error (MAE). The RMSE is squared error which is more sensitive to large deviation between forecasts and actuals. The MAE on the other hand mean absolute error, which is a more suitable measure. The MAE and RMSE values are lower for the hybrid LSTM-LA approach as compared to the FFNN and LSTM approaches, indicating that the hybrid LSTM-LA approach outperforms the standalone approaches LSTM and FFNN approach.

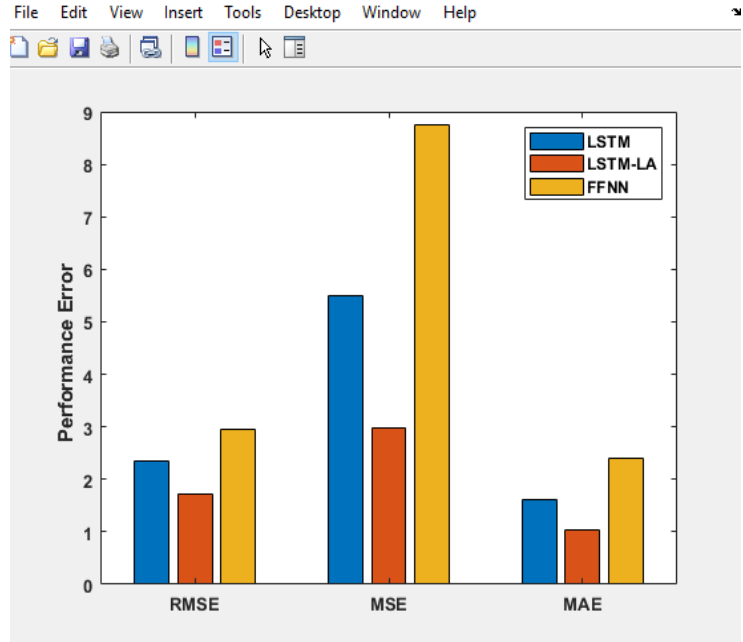

Figure 6: Plot for RMSE, MSE and MAE for the hybrid LSTM-LA, LSTM and FFNN models

The monthly forecast of GWL for the year 2019 using the hybrid LSTM-LA, LSTM and FFNN approaches are as presented (Figure 7). The graph shows increasing trend irrespective of season, because of inconsistency in rainfall.

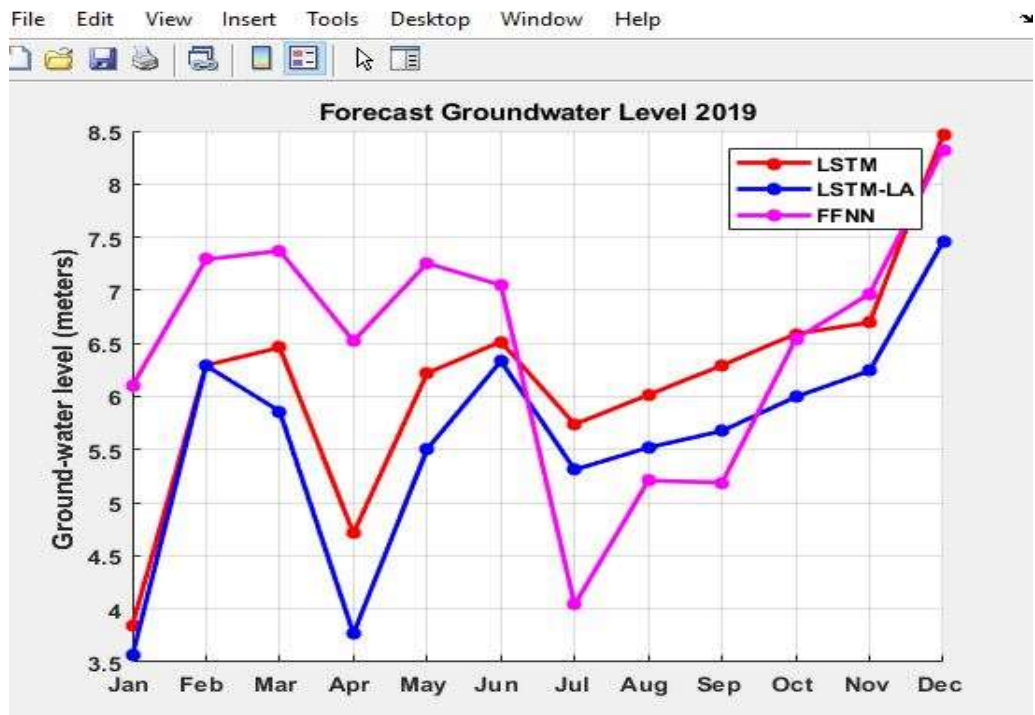

Figure 7: Plot for forecast groundwater level using LSTM-LA, LSTM and FFNN models

The time series plot below (Figure 8) shows the future GWL forecast using proposed hybrid LSTM-LA model. The model is trained using the data for a period of 216 months (18 years) starting from January 2000 to December 2017. The model is able to forecast future trend accurately up to a maximum of one year lead-time.

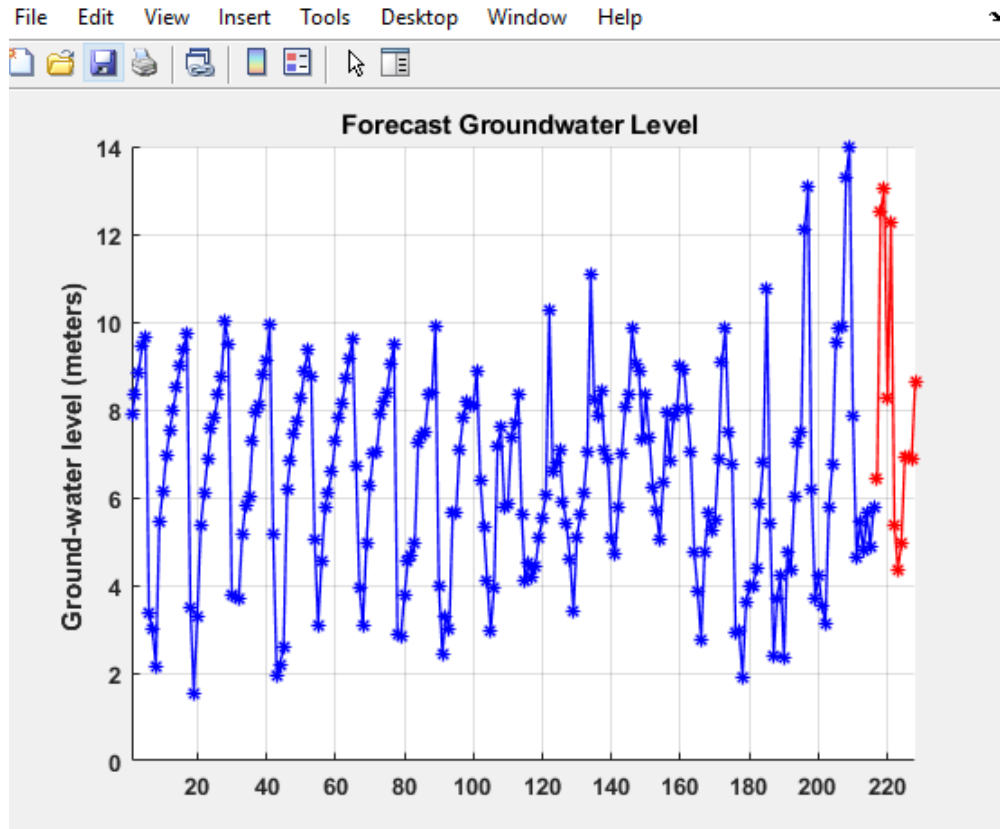

Figure 8: Plot for forecasting groundwater level

The cross validation resampling procedure is used to evaluate different machine learning algorithms on a limited data sample. Cross validation is primarily used to estimate the accuracy of machine learning algorithm on unseen data. In 5-fold cross validation, we partition the original training dataset into 5 equal subsets called folds. The accuracy of the machine learning algorithm is estimated by averaging the accuracies derived in all the 5 cases of cross validation. The box and whisker plot (Figure 9) shows the spread of the prediction accuracy scores across each validation fold for each algorithm.

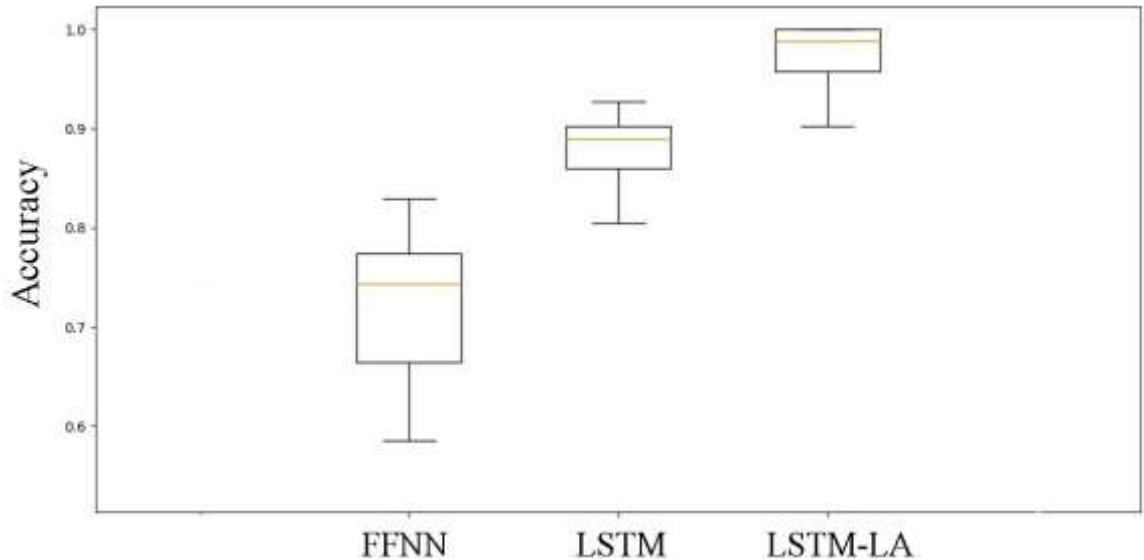

Figure 9: Box and whisker plot comparing the prediction accuracy of the models

The prediction accuracy, indicated by the median, for FFNN, LSTM and hybrid LSTM-LA approaches is 72%, 88% and 97.5%, respectively. The hybrid LSTM-LA based model is compared with traditional FFNN network structure based model. From the above plot in figure 9, it can be observed that LSTM-LA approach has higher accuracy compared to FFNN based model.

## 4. Conclusions

Scarcity of pure drinking water is the global problem. GWL gives useful information for assessing groundwater resource. The current study has developed a new hybrid metaheuristic approach using lion algorithm to optimise the weights of LSTM network for forecasting GWL. The precedent GWL and rainfall dataset from year 2000-2018 was accessed from government agencies. The observation well was located in a lateritic terrain in Udupi district, Karnataka, India. The results obtained from the propounded LSTM-LA model was compared with the basic FFNN and LSTM models. The FFNN model apprentice in randomised order, whereas feedback loops in LSTM enable to learn progressively. There are several drawbacks exploiting standalone LSTM network. It suffers from an unusual distribution of input variables in the test set compared to training data. Therefore, lion algorithm is used to optimise the weights of LSTM and developing LSTM-LA model. The lion algorithm looks for optimal point through different strategies by balancing exploration and exploitation. The hybrid LSTM-LA model is preferred over traditional FFNN and LSTM on its own, in terms of prediction accuracy and convergence rate. This research work concludes that GWL forecasting with systematically configured LSTM model surpasses traditional FFNN model with higher efficiency.

## Data Availability

The data used to support the findings of this study are available from the corresponding author upon request.

## Conflicts of Interest

There is no conflict of interest to report regarding the publication of this paper.

# Funding Statement

This research received no external funding.

# Acknowledgments

The authors thank Nagaraj Rao, Statistic Department, Udupi district, Government of Karnataka, Mr Niranjan and Mr Dinakar Shetty, Senior geologist, Udupi district, for providing valuable rainfall and GWL data for the study period. The authors are also grateful to Manipal Academy of Higher Education (MAHE) and Manipal Institute of Technology (MIT) for their support to this research work.